\documentclass[conference]{IEEEtran}
\usepackage{times}

\usepackage[numbers]{natbib}
\usepackage{multicol}
\usepackage[bookmarks=true]{hyperref}
\usepackage{amsmath} 
\usepackage{amssymb} 

\usepackage[utf8]{inputenc}
\usepackage{booktabs} 
\usepackage{multirow}
\usepackage{graphicx} 
\usepackage{makecell}
\usepackage{caption}

\begin{document}

\title{UniForce: A Unified Latent Force Model for Robot Manipulation with Diverse Tactile Sensors}




%
\author{\authorblockN{Zhuo Chen\authorrefmark{1},
Fei Ni\authorrefmark{2},
Kaiyao Luo\authorrefmark{1}, 
Zhiyuan Wu\authorrefmark{1}, 
Xuyang Zhang\authorrefmark{1},\\ 
Emmanouil Spyrakos-Papastavridis\authorrefmark{1}, 
Lorenzo Jamone\authorrefmark{2}, 
Nathan F. Lepora\authorrefmark{3}, 
Jiankang Deng\authorrefmark{2} and
Shan Luo\authorrefmark{1}}

\authorblockA{%
\authorrefmark{1}King's College London\quad
\authorrefmark{2}Imperial College London\quad
\authorrefmark{3}University College London\quad
\authorrefmark{4}University of Bristol}}



\twocolumn[{%
  \renewcommand\twocolumn[1][]{#1}%
  \maketitle
  \vspace{-5mm}

  \begin{center}
    \includegraphics[width=0.85\textwidth]{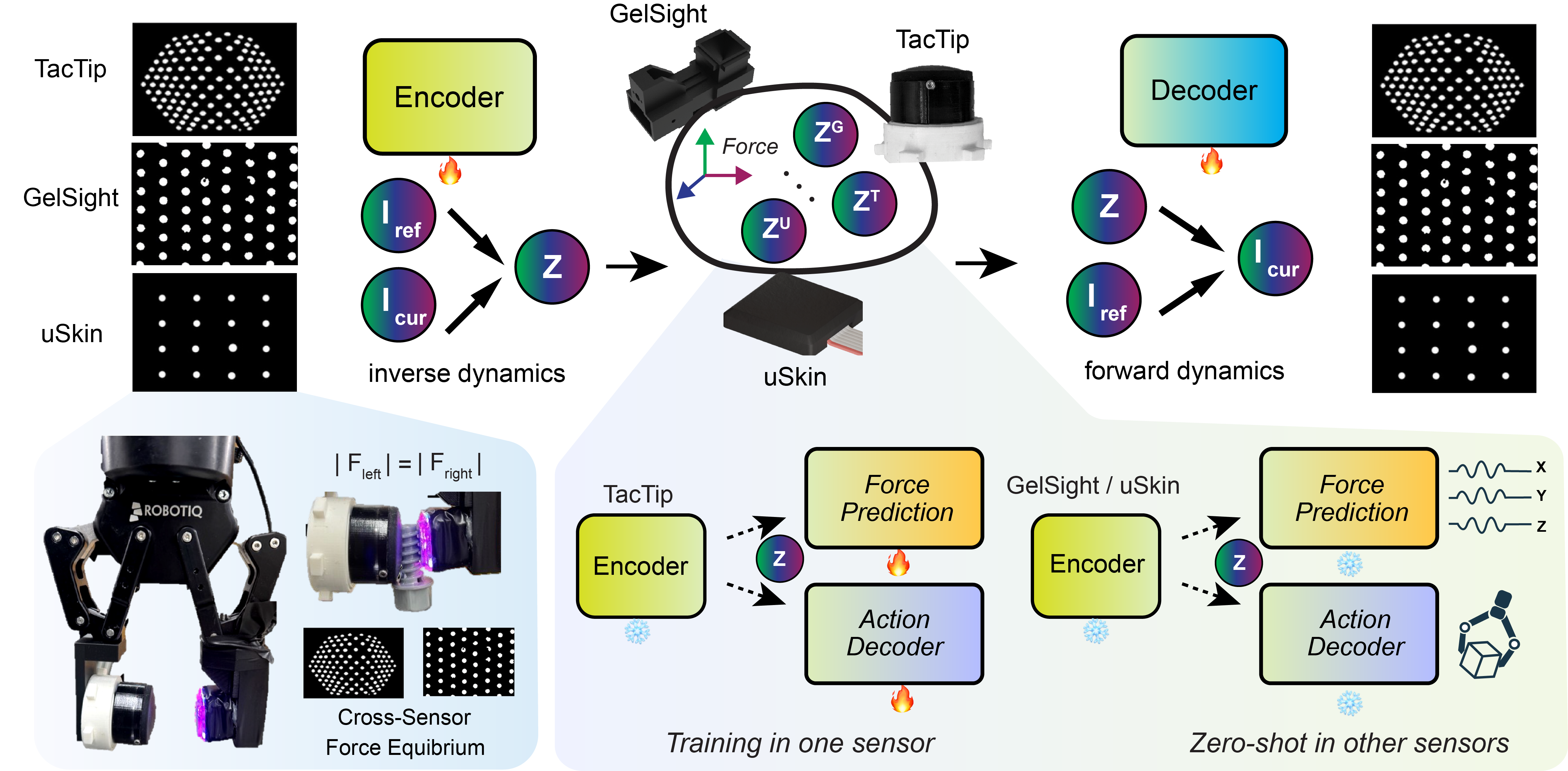}
    \captionof{figure}{%
    We introduce \textbf{UniForce}, a framework that learns a shared latent force model for robot manipulation across heterogeneous tactile sensors. Using paired data from force-equilibrium grasps, UniForce pretrains a universal encoder to extract physically meaningful latent forces from a unified marker-image representation \emph{without} explicit force labels. The pretrained encoder can then be plugged into downstream tasks by training specific head on data from a single sensor, enabling \textbf{zero-shot} transfer to other sensors without finetuning for cross-sensor force prediction and force-aware policy learning.
    }
    \label{fig:teaser}
  \end{center}
}]

\begin{abstract}
Force sensing is essential for dexterous robot manipulation, but scaling force-aware policy learning is hindered by the heterogeneity of tactile sensors. Differences in sensing principles (e.g., optical vs.\ magnetic), form factors, and materials typically require sensor-specific data collection, calibration, and model training, thereby limiting generalisability.
We propose \textbf{UniForce}, a novel unified tactile representation learning framework that learns a \emph{shared latent force space} across diverse tactile sensors.
UniForce reduces cross-sensor domain shift by jointly modeling inverse dynamics (image-to-force) and forward dynamics (force-to-image), constrained by force equilibrium and image reconstruction losses to produce force-grounded representations. To avoid reliance on expensive external force/torque (F/T) sensors, we exploit static equilibrium and collect force-paired data via direct sensor--object--sensor interactions, enabling cross-sensor alignment with contact force. The resulting universal tactile encoder can be plugged into downstream force-aware robot manipulation tasks with zero-shot transfer, without retraining or finetuning.
Extensive experiments on heterogeneous tactile sensors including GelSight, TacTip, and uSkin, demonstrate consistent improvements in force estimation over prior methods, and enable effective cross-sensor coordination in Vision-Tactile-Language-Action (VTLA) models for a robotic wiping task. Code and datasets will be released.
\end{abstract}

\IEEEpeerreviewmaketitle

\section{Introduction}
\label{sec:intro}

Human-level robotic manipulation benefits substantially from tactile feedback~\cite{luo2025tactile}, particularly force sensing~\cite{johansson2009coding}. However, learning force-aware models that \emph{transfer} across tactile sensors remains challenging.
The key challenge is sensor heterogeneity: tactile sensors differ in sensing principles (e.g., optical~\cite{GelSight1,Tactip}, magnetic~\cite{tomo2018uskin}, and capacitive~\cite{chen2022laser}), signal modalities (e.g., images vs.\ multi-channel signals), form factors (e.g., flat vs.\ curved surfaces), and mechanical properties (e.g., soft vs.\ hard elastomers).
As a result, force-aware policies and force prediction models trained on one sensor (e.g., TacTip~\cite{Tactip}) can degrade substantially when deployed on others such as GelSight~\cite{GelSight1} and uSkin~\cite{tomo2018uskin}. 
Moreover, collecting force labels and retraining/finetuning for each new sensor is costly and often impractical at scale, motivating a force-aware tactile representation that generalizes across heterogeneous sensors.

We argue that bridging this gap requires disentangling \textit{sensor-invariant contact force} from \textit{sensor-specific tactile signals}. A robot's own embodiment provides a scalable supervisory signal for this disentanglement: \textbf{cross-sensor force equilibrium}. During a quasi-static bilateral grasp with two fingers equipped with heterogeneous tactile sensors, physics constrains the opposing contact forces to match in magnitude, even though the resulting tactile measurements can differ substantially across sensors. Building on this prior, we present \textbf{UniForce}, a framework that learns a unified latent force model shared across diverse tactile sensors. Unlike prior work \cite{chen2025transforce,chen2026genforce,perla2024touch2touch,rodriguez2024touch2touch} that rely on location-paired data and image-to-image translation across sensors without explicitly grounding representations in contact dynamics, UniForce learns (i) an encoder with inverse dynamics that maps tactile images to force, and (ii) a decoder with forward dynamics that reconstructs tactile images from force (conditioned on the reference image), enabling self-supervised learning from force-paired tactile images. This physics-grounded objective encourages the learned latent force space to be invariant to sensor hardware, rather than entangling force with sensor-specific appearance, pixels or semantics as studied in prior approaches \cite{feng2025anytouch,xu2025unit,zhao2024transferable}.

To the best of our knowledge, UniForce is the first framework that learns a latent force space spanning both vision-based and non-vision-based tactile sensors. 
A recent work, GenForce~\cite{chen2026genforce}, achieves transferable force sensing across tactile sensors by transferring force labels from one sensor to others, but requires per-target diffusion-based image translation, force-model training, and material compensation. In contrast, UniForce learns a unified latent force space and a universal encoder in an end-to-end manner that can be plugged into force-aware policy learning and force prediction with zero-shot transfer, without sensor-specific retraining or finetuning. 

The contributions of this work are summarized as follows:
\begin{itemize}
    \item \textbf{A unified latent force model with a universal encoder}, UniForce, is proposed to learn shared force-aware representations across heterogeneous tactile sensors (e.g., vision-based and non-vision-based) by aligning them in a common latent force space.
    \item \textbf{A physics-grounded, label-free force-paired data collection pipeline} is introduced that exploits quasi-static force equilibrium to obtain force-paired tactile signals across heterogeneous tactile sensors without relying on external force/torque (F/T) sensors.
    \item \textbf{Zero-shot transfer for force-aware robot manipulation tasks} is achieved, in which the universal encoder transfers to downstream tasks across diverse tactile sensors, including force estimation and force-aware policy learning, without sensor-specific retraining or finetuning. 
\end{itemize}

\section{Related Works}
\label{sec:related}

\subsection{Heterogeneous Tactile Sensors}
Tactile sensors are featured by a soft skin and a sensing layer inspired from human skin \cite{luo2025tactile}. The sensing layer can rely on different transduction principles, including optical~\cite{GelSight1,Tactip}, magnetic~\cite{tomo2018uskin}, piezocapacitive~\cite{chen2022laser}, and piezoresistive~\cite{liu2024resistive} designs, each offering distinct trade-offs in resolution, sensitivity, and robustness.
In addition, the outer skin and form factor are often customized for specific manipulation tasks. For example, GelSight~\cite{GelSight1} is a flat, vision-based tactile sensor that outputs high-resolution RGB images and is commonly used for texture recognition and pose estimation. TacTip~\cite{Tactip} is a biomimetic vision-based tactile sensor with a curved surface and internal pins, making it sensitive to shear and contact forces. uSkin~\cite{tomo2018uskin} is a magnetic tactile sensor based on Hall-effect sensing that outputs multi-channel signals at high frequency, which is suitable for dynamic interactions. Due to the heterogeneity, force-aware policies are commonly trained for a specific sensor and do not directly transfer to other sensors or even to the same sensor with different skins~\cite{chen2026genforce}. Our work targets this fragmentation by learning a force-grounded representation that is invariant to sensor hardware. 

\subsection{Cross-Sensor Representation Learning}
Learning shared tactile representations and enabling cross-sensor transfer has recently gained attention in tactile robotics. Existing efforts can be broadly grouped into two lines. \textbf{(1) Representation learning within a sensor family.}
Several methods learn a shared latent space for vision-based tactile sensors, particularly the GelSight family. AnyTouch~\cite{feng2025anytouch} learns unified representations across static and dynamic tactile observations. T3~\cite{zhao2024transferable} uses a shared trunk with sensor-specific encoders and task-specific decoders. UniT~\cite{xu2025unit} learns tactile representations with a VQ-VAE-style objective. These approaches demonstrate strong transfer within a narrow hardware regime, but they typically assume similar sensing modality and do not explicitly enforce that the latent space corresponds to a physical variable (e.g., contact force), which limits transferability to non-vision-based sensors and deployment on robot skill learning.  \textbf{(2) Cross-sensor transfer via self-supervision or translation.}
Another line performs cross-sensor transfer using self-supervised learning or image-to-image translation between tactile domains~\cite{perla2024touch2touch,rodriguez2024touch2touch}. However, many of these methods are trained for specific sensor pairs and do not scale well as the number of sensors grows, especially across different transduction principles. GenForce~\cite{chen2026genforce} is closely related in that it targets transferable force sensing among heterogeneous sensors, but it relies on force labels with per-sensor translation and training. In contrast, UniForce aligns heterogeneous sensors through a shared latent force space grounded by cross-sensor force equilibrium, aiming to support plug-and-play zero-shot use across modalities.

\subsection{Force-Aware Robot Manipulation}
Force signals and tactile observations are increasingly integrated into policy learning, such as Vision-Language-Action (VLA) models~\cite{pi05}. A common strategy is to use pretrained tactile encoders to produce tactile tokens for high-level policy conditioning, as in OmniVTLA~\cite{cheng2025omnivtla} and TactileVLA~\cite{huang2025tactilevla}, often building on vision-based tactile representation models (e.g., AnyTouch~\cite{feng2025anytouch}). Another strategy fuses tactile information into both high-level perception and low-level action decoding to improve robustness \cite{bi2025vlatouch}. However, existing tactile encoders are trained to be grounded in language semantics, which limit their usefulness in force-centric manipulation tasks and for non-vision-based tactile sensors. ForceVLA~\cite{yu2025forcevla} incorporates 6D wrist force/torque sensing via a mixture-of-experts design, but wrist F/T measurements cannot directly resolve contact state variables such as slippage or distributed grasp forces. On  the other hand, directly using force estimation from tactile sensing faces practical challenges, most notably the need for expensive paired force-tactile data and force model training for each sensor ~\cite{chen2025transforce}. UniForce complements these directions by providing a universal, force-grounded encoder that can serve as a shared latent space across heterogeneous tactile sensors without the need for force--labels. 

\section{Methodology}
\label{sec:methodology}

We propose \textbf{UniForce}, a framework that learns a shared latent force representation across heterogeneous tactile sensors by exploiting quasi-static force equilibrium. The key observation is that although tactile measurements vary drastically across sensors due to differences in sensing principles, form factors, and skin materials, the underlying contact interaction in a bilateral, quasi-static grasp imposes a physical constraint: the two fingertips experience approximate equal-and-opposite contact forces. UniForce turns this constraint into implicit supervision to align heterogeneous tactile observations in a common latent force space through inverse--forward dynamics.

\begin{figure*}[t]
    \centering
    \includegraphics[width=0.85\linewidth]{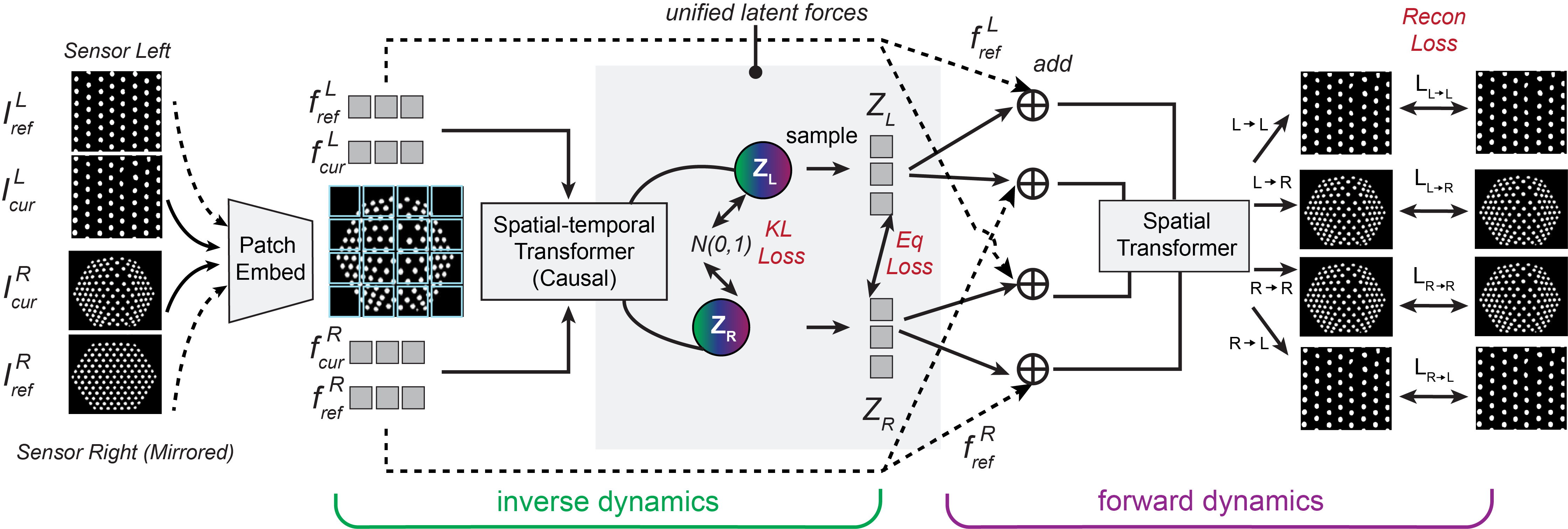}
    \caption{\textbf{Unified force-aware tactile representation learning.}
    In \textbf{the inverse-dynamics stage}, the encoder takes paired reference/contact observations and infers a patch-wise latent force map; an equilibrium loss aligns left/right latent forces, while a KL term regularizes the posterior. In \textbf{the forward-dynamics stage}, the decoder reconstructs the observation by conditioning on the sensor-specific reference image and the latent force, enabling both self-reconstruction and cross-sensor reconstruction. We mirror right-finger marker images to match the coordinates of the left finger.}
    \vspace{-7pt}
    \label{fig:model}
\end{figure*}

\subsection{Problem Formulation}
Let $\mathcal{S} = \{s_1, \dots, s_N\}$ denote a set of heterogeneous tactile sensors. Each sensor may have a distinct taxel layout or physical marker pattern.
Following prior work~\cite{chen2026genforce}, we canonicalize each sensor's raw signal into a unified 2D marker image representation, allowing a single image-based architecture to be applied across sensing modalities. We define a tactile observation as $\mathbf{x} = \langle I_{\text{ref}}, I_{\text{cur}} \rangle$, where $I_{\text{ref}}$ is a reference frame at the undeformed state ($t=0$) and $I_{\text{cur}}$ is a contacted frame at time $t=T$ capturing elastomer deformation under contact. Consider a synchronized sensor pair $\{s_L, s_R\}$ grasping an object with symmetric morphology. Although $\mathbf{x}_L$ and $\mathbf{x}_R$ can differ substantially due to domain shift (e.g., marker density and material property), quasi-static equilibrium constrains the interaction force at the two fingertips to be approximately equal and opposite. For simplicity, we denote this constraint as $F_L \approx F_R$ in magnitude, consistent with Fig.~\ref{fig:teaser}. Our goal is to learn a sensor-invariant encoder $\mathcal{E}_\theta$ that maps sensor-specific observations to a shared latent force space. We represent contact force as a patch-wise latent force map $\mathbf{z} \in \mathbb{R}^{N_p \times 6}$, where $N_p$ is the number of spatial patches and each patch corresponds to a 6D latent force vector. We formulate cross-sensor alignment as matching the encoder posteriors inferred from the left and right observations:
\begin{equation}
    q_\theta(\mathbf{z} \mid \mathbf{x}_L) \approx q_\theta(\mathbf{z} \mid \mathbf{x}_R)
\end{equation}

\subsection{Architecture: The UniForce Model}
UniForce is a conditional variational autoencoder (CVAE) that disentangles contact physics from sensor-specific tactile signals (Fig. \ref{fig:model}).
The encoder approximates inverse dynamics (tactile observation $\to$ force latent), and the decoder model's forward dynamics (force latent $\to$ contacted tactile observation) conditioned on the sensor's reference observation.

\subsubsection{Encoder -- Inverse Dynamics}
\label{subsec:encoder}
Given $\mathbf{x} = \langle I_{\text{ref}}, I_{\text{cur}} \rangle \in \mathbb{R}^{2 \times 3 \times H \times W}$, we patchify the two frames into tokens $\mathbf{V}_0 \in \mathbb{R}^{T \times N_p \times D}$, where $T=2$, $N_p$ is the number of patches, and $D$ is the embedding dimension.
We inject sinusoidal positional encodings to preserve the temporal order from reference to contact.
To extract deformation features, we adopt a Causal Spatiotemporal Transformer~\cite{ye2024latentaction} with factorized attention for efficiency and to enforce causality:

\begin{itemize}

    \item \textbf{Spatial attention} operates within each frame to extract local marker structures via multi-head self-attention, followed by LayerNorm and residual connections.
    \item \textbf{Causal temporal attention} operates along the temporal axis for spatially aligned tokens. A causal mask $M_t \in \mathbb{R}^{T \times T}$ allows the contacted frame ($t=T$) to attend to the reference frame ($t=0$), while preventing the reference frame from attending to current contact.
\end{itemize}

\subsubsection{Probabilistic Force Manifold}
The encoder outputs $\mathbf{h}_{\text{last}} \in \mathbb{R}^{N_p \times D}$, which is projected to a diagonal-Gaussian posterior over the patch-wise force map:
\begin{equation}
    [\boldsymbol{\mu}, \log\boldsymbol{\sigma}^2]
    = \text{Linear}(\text{LayerNorm}(\mathbf{h}_{\text{last}}))
    \in \mathbb{R}^{N_p \times 12}
\end{equation}
During training, we sample $\mathbf{z} = \boldsymbol{\mu} + \boldsymbol{\sigma} \odot \boldsymbol{\epsilon}$ with $\boldsymbol{\epsilon} \sim \mathcal{N}(\mathbf{0}, \mathbf{I})$. During inference, we use $\mathbf{z}=\boldsymbol{\mu}$.

\subsubsection{Decoder -- Physics-Conditioned Forward Dynamics}
\label{subsec:decoder}
The decoder $\mathcal{D}_\psi$ reconstructs the contacted image $\hat{I}_{\text{cur}}$ by conditioning on the reference image $I_{\text{ref}}$, which implicitly provides sensor-specific appearance (e.g., marker/taxel distribution).
We fuse latent force with the reference embedding via additive injection:
\begin{equation}
    \mathbf{V}_{\text{dec}} = \text{PatchEmbed}(I_{\text{ref}}) + \text{Linear}_{\text{proj}}(\mathbf{z})
\end{equation}
A spatial transformer then predicts the pixel-wise reconstruction $\hat{I}_{\text{cur}}$ subject to image reconstruction loss.

\subsection{Cross-Sensor Force Equilibrium Training}
\label{subsec:training}
We train UniForce without ground-truth force labels using a three-branch objective (Fig. \ref{fig:model}) over a paired grasp $(L,R)$.
Let $\mathbf{x}_i=\langle I_{\text{ref}}^i, I_{\text{cur}}^i\rangle$ for $i\in\{L,R\}$, and let $\mathbf{z}_i \sim q_\theta(\mathbf{z}\mid \mathbf{x}_i)$.
For each ordered pair $(i,j)\in\{L,R\}\times\{L,R\}$, we decode
\begin{equation}
    \hat{I}_{\text{cur}}^{i\to j} = \mathcal{D}_\psi(I_{\text{ref}}^j, \mathbf{z}_i)
\end{equation}
where $i=j$ corresponds to self-reconstruction and $i\neq j$ corresponds to cross-sensor reconstruction.
Under quasi-static equilibrium, the supervision target for $\hat{I}_{\text{cur}}^{i\to j}$ is $I_{\text{cur}}^j$, which encourages $\mathbf{z}_i$ to retain force-relevant content while discarding sensor-specific appearance.

\subsubsection{Objective Function}
The total loss combines reconstruction, KL regularization, and equilibrium consistency:
\begin{equation}
    \mathcal{L}_{total} = \mathcal{L}_{recon} + \lambda_{kl}\mathcal{L}_{KL} + \lambda_{eq}\mathcal{L}_{eq}
\end{equation}

\noindent\textbf{Reconstruction loss ($\mathcal{L}_{recon}$):}
We combine an $\ell_1$ loss with LPIPS~\cite{zhang2018LPIPS} to capture marker deformation in four branches:
\begin{equation}
    \mathcal{L}_{recon}
    = \sum_{i,j \in \{L,R\}}
    \left(
    \left\| \hat{I}_{\text{cur}}^{i\to j} - I_{\text{cur}}^{j} \right\|_1
    + \lambda_{lpips}\,\text{LPIPS}\!\left(\hat{I}_{\text{cur}}^{i\to j}, I_{\text{cur}}^{j}\right)
    \right)
\end{equation}

\noindent\textbf{KL divergence ($\mathcal{L}_{KL}$):}
We regularize $q_\theta(\mathbf{z}\mid \mathbf{x})$ towards $\mathcal{N}(\mathbf{0}, \mathbf{I})$ as in standard VAEs.

\noindent\textbf{Equilibrium loss ($\mathcal{L}_{eq}$):}
We enforce patch-wise consistency between the left and right latent force maps:
\begin{equation}
    \mathcal{L}_{eq} = \frac{1}{N_p}\left\| \mathbf{z}_L - \mathbf{z}_R \right\|_2^2
\end{equation}

\section{Force-Equilibrium Data Collection}
\label{sec:implementation}

\subsection{Setup}

\begin{figure}[htbp]
    \centering
    \includegraphics[width=0.7\linewidth]{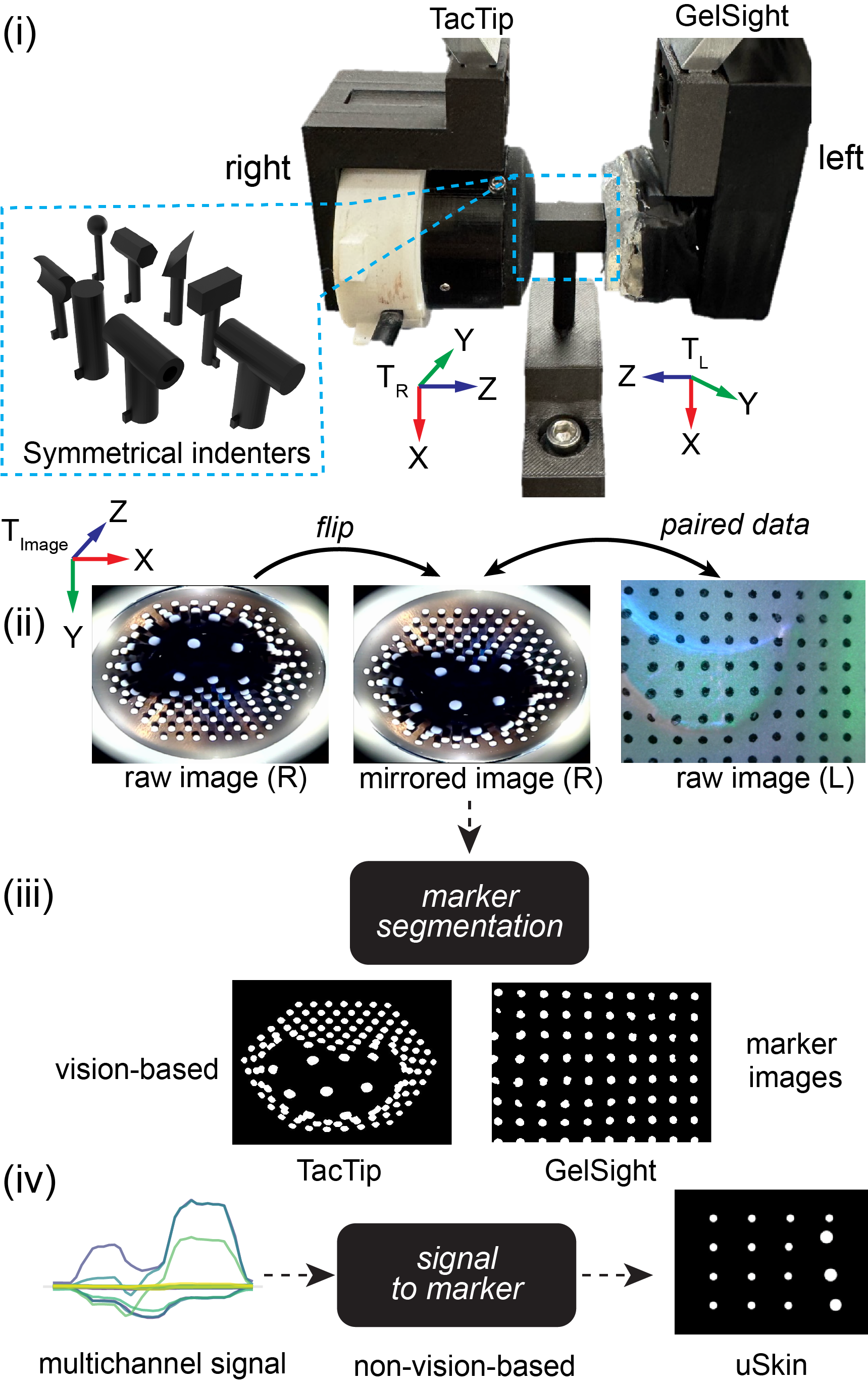}
    \caption{\textbf{Paired data collection via quasi-static force equilibrium.}
    (i) Data collection setup;
    (ii) Coordinate alignment by mirroring.
    (iii--iv) Canonicalizing heterogeneous raw signals referring to ~\cite{chen2026genforce}.}
    \vspace{-7pt}
    \label{fig:data_collection}
\end{figure}

\subsubsection{Hardware}
As shown in Fig.~\ref{fig:data_collection}, our setup consists of a Franka FR3 robot arm with a Franka Hand, symmetric indenters, and tactile sensors mounted on the two fingers.

We use three tactile sensors:
\begin{itemize}
    \item \textbf{GelSight 2017 (vision-based) \cite{GelSight1}.}
    We fabricate a GelSight sensor in-house using a Logitech C270 webcam (720p, 30\,fps) inside a 3D-printed housing with RGB LEDs.
    The silicone elastomer is printed with a square marker pattern (80 markers). The sensing area is $25 \times 25 \times 4$\,mm, with a maximum indentation depth of 1.5\,mm. 
    \item \textbf{TacTip (vision-based)~\cite{zhang2024tacpalm}.}
    TacTip is built with a fisheye camera (ELP USB camera, 1080p, 30\,fps) inside a 3D-printed shell with white LEDs.
    The outer skin is Agilus30, backed by a transparent gel that is exceptionally soft.
    The sensor contains 127 internal markers, has a curvature radius of 41.5\,mm and a diameter of 40\,mm, and allows up to 5\,mm indentation.
    \item \textbf{uSkin (magnetic-based)~\cite{tomo2018uskin}.}
    We use the XELA Robotics uSkin (patch version) with a $4\times4$ taxel array (16 taxels). Each taxel provides 3-axis deformation signals.
    The sensor size is $24.6 \times 22.6 \times 5$\,mm and it outputs multi-channel signals at 500\,Hz.
\end{itemize}

We design a set of symmetric, double-sided indenters with eight shapes (e.g., square, triangle, curved, and spherical primitives). The indenter size is designed to remain within each sensor's effective sensing area. The indenter fixture is mounted on a base rigidly attached to a metallic breadboard. 

\subsubsection{Data Collection Protocol}
We use TacTip as an \emph{anchor} sensor fixed on the right finger, while swapping the left finger sensor between GelSight and uSkin. This yields two paired datasets, $\langle \text{GelSight}, \text{TacTip}\rangle$ and $\langle \text{uSkin}, \text{TacTip}\rangle$, which we refer to as the \textit{UniForce-pair} dataset. 

The robot runs in cartesian-space impedance control. We set a high stiffness along the gripper closing direction to reduce unintended lateral motion during contact, and we start from a gentle initial grasp where both sensors observe a small indentation.
A human operator then slowly drags or rotates the robot hand to induce tangential deformation while gradually increasing the indentation depth. These actions are performed at low speed to approximate a quasi-static state in which force equilibrium between the two fingertips is expected to hold. For each indenter, we record 1{,}000 frames, resulting in approximately 8{,}000 frames per sensor pair. Data collection takes less than 30 minutes per sensor pair.

\subsubsection{Data Preprocessing}
\textbf{1). Coordinate alignment by mirroring.}
When using the image frame as the reference coordinate system, the y-axes of the left and right fingertips are mirrored (Fig.~\ref{fig:data_collection}). To align the coordinate systems, we horizontally flip the right-finger tactile images before forming force-equilibrium pairs, so that contact geometry and force direction are represented consistently across fingers. \textbf{2). Canonicalizing heterogeneous raw signals.} Raw RGB tactile images can contain illumination/background variations, and uSkin outputs multi-channel time-series signals. Following GenForce~\cite{chen2026genforce}, we convert both vision-based tactile images and non-vision tactile signals into a unified binary marker-image representation using either image segmentation (for vision-based sensors) or a signal-to-marker rendering process (for uSkin). This preprocessing runs at over 29.6\,Hz in our implementation.
All marker images are resized to $640 \times 480$ for training. 

\begin{figure}[t]
    \centering
    \includegraphics[width=0.9\linewidth]{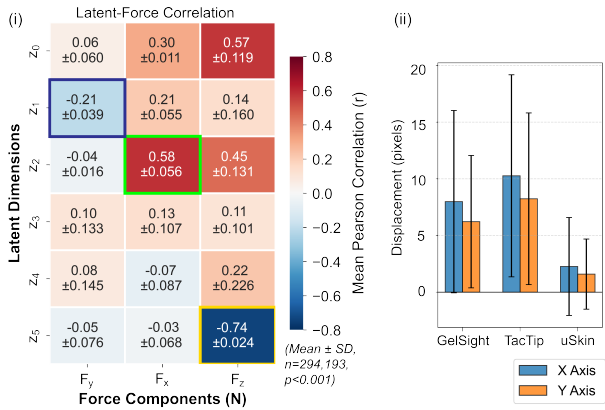}
    \caption{(i) Heatmap of mean Pearson correlation \(r\) (\(\pm\) SD) between latent dimensions \(z_0\)–\(z_5\) and force components \(F_y, F_x, F_z\) (N). Total images \(n=294{,}193\) across three sensors; \(p<0.001\) (ii) Absolute mean displacement of marker in x-axis and y-axis from \textit{UniForce-pair} dataset.}
    \vspace{-7pt}
    \label{fig:coorelation}
\end{figure}

\begin{figure}[t]
    \centering
    \includegraphics[width=1\linewidth]{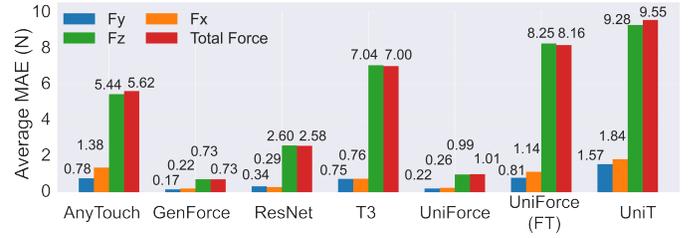}
    \caption{\textbf{Average zero-shot force prediction errors (MAE).} We compare MAE across six heterogeneous sensor-pairs. FT denotes full-training.}
    \vspace{-7pt}
    \label{fig:force_average}
\end{figure}

\begin{figure*}[t]
    \centering
    \includegraphics[width=0.9\linewidth]{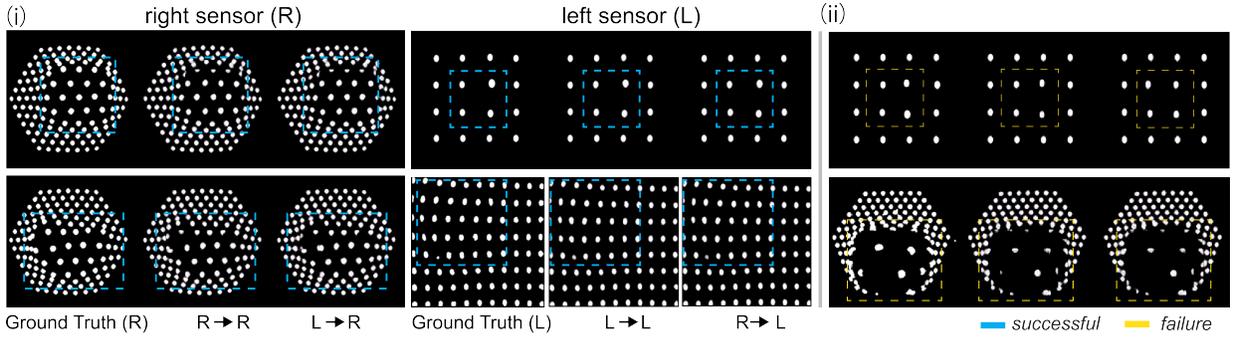}
    \caption{\textbf{Cross-sensor tactile image reconstruction (GelSight--TacTip and uSkin--TacTip).}
    TacTip on the right finger serves as the anchor sensor. The inferred latent forces drive four reconstruction branches (self-reconstruction and cross-sensor reconstruction) through the forward-dynamics decoder. A single UniForce model is trained jointly on all three sensors.}
    \vspace{-7pt}
    \label{fig:image_recons}
\end{figure*}

\begin{table*}[t]
\centering
\scriptsize
\setlength{\tabcolsep}{1.5pt}
\caption{Zero-shot force prediction comparison using $R^2$ on \textbf{unseen} objects.
Abbreviations: \textbf{G}: GelSight, \textbf{T}: TacTip, \textbf{U}: uSkin. G$\to$T denotes transfer from G to T. Higher is better ($\uparrow$). }
\label{tab:force_R2}

\resizebox{\textwidth}{!}{%
\begin{tabular}{l ccc ccc ccc ccc ccc ccc}
\toprule
\multirow{2}{*}{\textbf{Model}} &
\multicolumn{3}{c}{\textbf{G $\to$ T}} & \multicolumn{3}{c}{\textbf{G $\to$ U}} &
\multicolumn{3}{c}{\textbf{T $\to$ G}} & \multicolumn{3}{c}{\textbf{T $\to$ U}} &
\multicolumn{3}{c}{\textbf{U $\to$ G}} & \multicolumn{3}{c}{\textbf{U $\to$ T}} \\
\cmidrule(lr){2-4} \cmidrule(lr){5-7} \cmidrule(lr){8-10}
\cmidrule(lr){11-13} \cmidrule(lr){14-16} \cmidrule(lr){17-19}
& $F_x\!\uparrow$ & $F_y\!\uparrow$ & $F_z\!\uparrow$ & $F_x\!\uparrow$ & $F_y\!\uparrow$ & $F_z\!\uparrow$ & $F_x\!\uparrow$ & $F_y\!\uparrow$ & $F_z\!\uparrow$ & $F_x\!\uparrow$ & $F_y\!\uparrow$ & $F_z\!\uparrow$ & $F_x\!\uparrow$ & $F_y\!\uparrow$ & $F_z\!\uparrow$ & $F_x\!\uparrow$ & $F_y\!\uparrow$ & $F_z\!\uparrow$ \\
\midrule
ResNet & -0.08 & -1.66 & 0.01 & -0.01 & -0.71 & 0.04 & 0.02 & -0.22 & -0.26 & 0.0 & -0.01 & -0.09 & 0.03 & -0.14 & -2.42 & -0.25 & -0.12 & -9.34 \\
AnyTouch & -4.11 & -0.31 & -5.41 & -1.97 & -0.48 & -12.23 & -0.94 & 0.23 & -0.07 & -0.91 & -3.91 & -6.19 & -38.84 & -13.3 & 0.23 & -43.04 & -34.82 & -44.27 \\
T3 & -0.2 & -0.23 & -3.49 & -0.03 & -0.2 & 0.17 & 0.3 & 0.2 & 0.1 & -0.21 & -0.38 & -0.5 & -15.93 & -46.56 & -59.85 & -22.86 & -8.41 & -130.17 \\
UniT & -1.47 & -1.18 & -33.74 & -12.53 & -0.64 & -0.06 & 0.13 & 0.13 & 0.31 & -0.53 & -0.5 & -6.11 & -57.73 & -122.61 & -79.34 & -105.75 & -123.13 & -157.72 \\
UniForce (FT) & -0.48 & -0.99 & -14.52 & -1.43 & -0.04 & -4.65 & -0.49 & -2.6 & -0.85 & -5.72 & -0.77 & -1.69 & -5.9 & -21.68 & -37.22 & -61.84 & -33.62 & -155.04 \\
\textbf{UniForce} & \textbf{0.1} & \textbf{0.17} & \textbf{0.59} & \textbf{0.4} & \textbf{0.29} & \textbf{0.47} & \textbf{0.77} & \textbf{0.62} & \textbf{0.83} & \textbf{0.35} & \textbf{0.3} & \textbf{0.39} & \textbf{-0.23} & \textbf{-0.51} & \textbf{0.17} & \textbf{-2.36} & \textbf{-0.55} & \textbf{-1.51} \\
\midrule
GenForce* & 0.38 & 0.3 & 0.63 & 0.59 & 0.69 & 0.79 & 0.48 & 0.49 & 0.67 & 0.26 & 0.41 & 0.55 & 0.58 & 0.62 & 0.72 & 0.3 & 0.34 & 0.61 \\
\bottomrule
\end{tabular}%
}
\vspace{2pt}
{\footnotesize *GenForce transfers and trains force prediction models per target sensor (with additional per-sensor processing), and is non-zero-shot method.} 
\end{table*}

\section{Experiments \& Results}
\label{sec:experiments}

We evaluate UniForce to answer four questions:
\begin{itemize}
  \item[\textbf{Q1}] Does the learned latent space encode contact force in the inverse-dynamics encoder?
  \item[\textbf{Q2}] Can the decoder use the latent force to reconstruct deformed tactile observations across heterogeneous sensors (forward dynamics)?
  \item[\textbf{Q3}] How does UniForce perform in terms of zero-shot force estimation compared with prior tactile representation learning methods?
  \item[\textbf{Q4}] Can UniForce be integrated into skill learning (e.g., Vision--Language--Action models) to improve force-aware manipulation with heterogeneous tactile sensing?
\end{itemize}

\subsection{Learned Latent Force Space (Q1)}
To answer \textbf{Q1}, we freeze the UniForce encoder and analyze whether the inferred latent dimensions (\(z_0\)–\(z_5\), with \(\mathbf{z}\in\mathbb{R}^{6}\)) correlate with ground-truth force labels. 

\textbf{Dataset.}
We use the \textbf{GenForce-Hetero} dataset~\cite{chen2026genforce}, which contains tactile sequences paired with force labels for TacTip (108{,}153 images), GelSight (106{,}278 images), and uSkin (79{,}762 images).
Forces are applied using 18 indenters with normal forces in the range \([0,12]\) N and shear forces in the range \([-3,3]\) N.
Note that the GelSight in GenForce-Hetero uses a diamond-like marker pattern (GelSight-D), which differs from our in-house GelSight sensor; TacTip and uSkin match our hardware. 

\textbf{Metric.}
We compute Pearson correlations between each latent dimension \(z_i\) and each force component \((F_x,F_y,F_z)\), reporting mean \(\pm\) SD (all \(p<0.001\)).
We obtain \(z_i\) by averaging over \(N_p=196\) spatial patches. In total, we analyze \(n=294{,}193\) images across the three sensors.

\textbf{Normal force emerges as a stable latent factor.}
Fig.~\ref{fig:coorelation}i shows that \(F_z\) is strongly associated with a single latent dimension \(z_5\) across sensors (\(r=-0.74\pm0.024\)), corresponding to linear predictability \(R^2=0.552\) from \(z_5\) alone (TacTip: \(r=-0.768\), \(R^2=0.590\); GelSight: \(r=-0.748\), \(R^2=0.559\); uSkin: \(r=-0.711\), \(R^2=0.506\)). \textbf{Shear force is encoded moderately.}
For lateral shear, \(F_x\) correlates most with \(z_2\) (\(r=0.58\pm0.056\), \(R^2=0.342\)).
In contrast, \(F_y\) is weakly encoded (mean \(|r|\approx 0.21\), \(R^2=0.047\)), and the best-correlated dimension varies across sensors.
We attribute this primarily to dataset composition: deformation is dominated by grasping-induced normal force (\(F_z\)) and a downward tangential component (\(F_x\)) due to hand weight, while \(F_y\) is less excited (Fig.~\ref{fig:coorelation}ii), making it harder to infer from tactile observations.
Overall, these correlations support that UniForce recovers physically meaningful latent factors from self-supervised training. 

\subsection{Cross-Sensor Image Reconstruction (Q2)}
UniForce is trained to reconstruct the contacted tactile observation \(I_{\text{cur}}\) from the reference observation \(I_{\text{ref}}\) and the inferred latent force, which serves as a forward-dynamics surrogate.
A distinctive capability is \emph{cross-sensor reconstruction}: using a latent force inferred from one sensor to reconstruct the contacted observation of another sensor when the grasp is in quasi-static equilibrium.

\begin{figure*}[htp]
    \centering
    \includegraphics[width=0.8\linewidth]{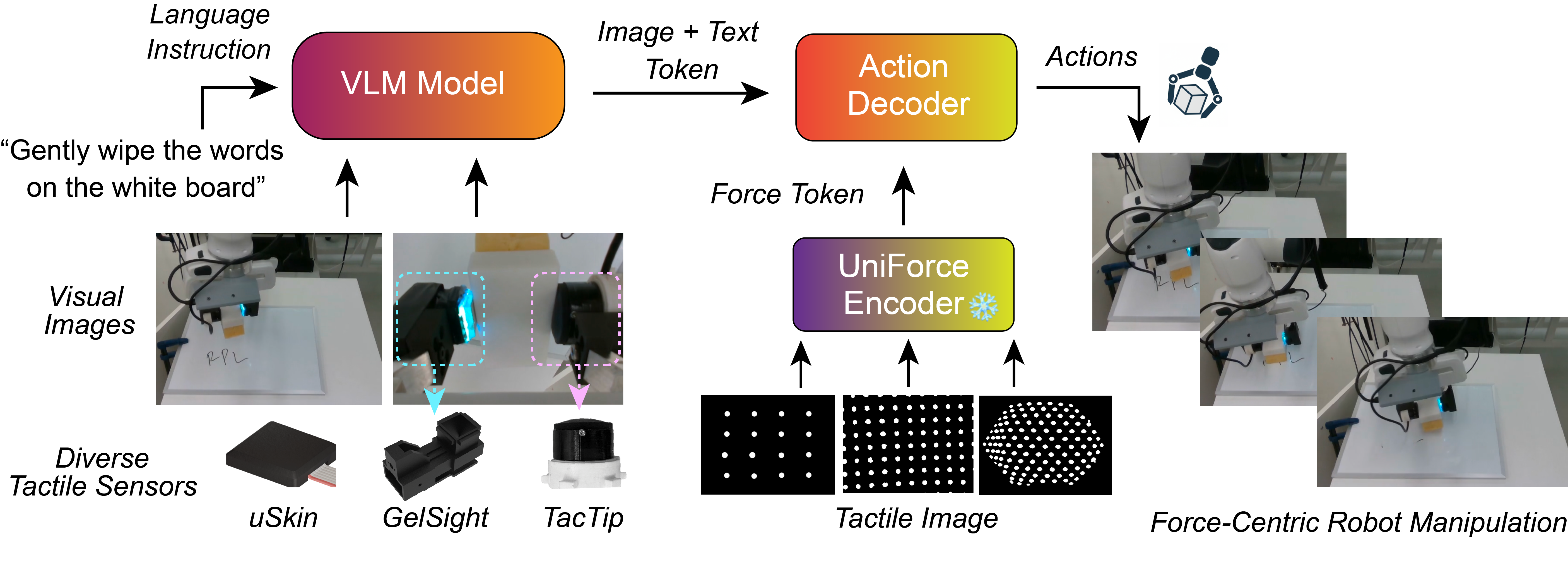}
    \caption{\textbf{Plugging Uniforce in Vision-Language-Action model}. Heterogeneous tactile inputs unified as force-grounded tokens via UniForce for force-aware robot manipulation task, i.e., whiteboard wiping.}
    \vspace{-7pt}
    \label{fig:vtla}
\end{figure*}

As shown in Fig.~\ref{fig:image_recons}, UniForce reconstructs both self-sensor and cross-sensor tactile responses. For example, latent force inferred from TacTip can drive reconstructions for GelSight and uSkin (conditioned on their reference observations), and latent force inferred from GelSight/uSkin can drive TacTip reconstructions.

While identical contact forces can yield different marker displacements due to contact location and indenter geometry, the reconstructions preserve consistent contact location and geometry with ground truth, enabled by the patch-wise latent design (e.g., \(16\times16\) patches). We also observe failure cases. For instance, uSkin cross-reconstructions can show inconsistent marker displacement, and TacTip reconstructions conditioned on GelSight can be incomplete under large deformations. These cases highlight the challenge of accurately rendering fine-grained contact patterns across sensors with sparse signals or large deformation ranges.
Nevertheless, successful reconstructions indicate that the decoder has learned a reference-conditioned mapping consistent with the intended forward model: \(I_{\text{cur}} \approx \mathcal{D}(I_{\text{ref}}, \mathbf{z})\). 

\subsection{Zero-Shot Force Estimation (Q3)}
To answer \textbf{Q3}, we freeze the pretrained UniForce encoder and attach a trainable force prediction head.
We train the head using one source sensor from the \textbf{GenForce-Hetero} dataset~\cite{chen2026genforce}  (GelSight-D (G), TacTip (T), or uSkin (U)), and then evaluate on the other sensors without any additional finetuning. In the training stage, we use the seen objects (12 types) from the dataset. While in the test stage, we use the unseen objects (6 types).

\textbf{Baselines.}
We compare against pretrained tactile representation models (frozen using official checkpoints when available) and standard backbones trained end-to-end.
For a fair comparison, we attach the same force prediction head (ConvGRU+MLP) to each frozen encoder and predict force from tactile sequences, following prior results showing benefits over single-frame prediction~\cite{chen2025transforce}.
\begin{itemize}
    \item \textbf{AnyTouch~\cite{feng2025anytouch}:} Representation learning for the GelSight family trained with both static and dynamic data.
    \item \textbf{T3~\cite{zhao2024transferable}:} Shared trunk with sensor-specific encoders and task-specific decoders for the GelSight family.
    \item \textbf{UniT~\cite{xu2025unit}:} VQ-VAE-style tactile representation trained on GelSight Mini with markers.
    \item \textbf{ResNet~\cite{he2016deep}:} A common end-to-end backbone for force prediction (full training).
    \item \textbf{UniForce (FT):} Fully trained force model without freezing the UniForce encoder.
    \item \textbf{GenForce~\cite{chen2026genforce}:} Transferable force sensing with per-sensor translation and per-sensor training; we report it as a \textbf{non-zero-shot reference}. 
\end{itemize}

\textbf{Metrics.}
We report coefficients of determination (\(R^2\)) and mean absolute errors (MAE) between predicted and ground-truth forces. Note that negative \(R^2\) indicates performance worse than predicting the mean force. 

Table~\ref{tab:force_R2} and Fig.~\ref{fig:force_average} show that UniForce improves zero-shot transfer over the baselines across all sensor-pair settings. Even compared with the non-zero-shot method, when transferring from TacTip to GelSight (T\(\rightarrow\)G), UniForce attains \(R^2=\{0.77,0.62,0.83\}\) for \((F_x,F_y,F_z)\) with better performance compared with GenForce's \(\{0.48,0.49,0.67\}\). In addition, it achieves MAE close to GenForce on average across the six transfer settings (UniForce: \(0.22\) N/\(0.26\) N/\(0.99\) N vs.\ GenForce: \(0.17\) N/\(0.22\) N/\(0.73\) N for \(F_x/F_y/F_z\)). It may be attributed to UniForce’s end-to-end training with force-ground latent space, which reduces information loss and image-translation noise, compared with GenForce’s three-stage pipeline of image-to-image translation, force-prediction model training, and material compensation. We also observe weaker transfer when using uSkin as the source sensor (U\(\rightarrow\)G and U\(\rightarrow\)T), consistent with the information bottleneck when transferring from a lower-resolution tactile signal to denser marker images similar to GenForce. Nevertheless, the strong improvements in force errors and $R^2$ values compared with baselines validates the effectiveness of UniForce in zero-shot transfer of force estimation across heterogeneous sensors.

\subsection{Zero-Shot Force-Aware Policy Transfer (Q4)}
To answer \textbf{Q4}, we integrate the pretrained UniForce encoder into $\pi_{0.5}$ \cite{pi05}, a Vision--Language--Action model, for force-aware skill learning, i.e., whiteboard wiping (Fig.~\ref{fig:vtla}).
The UniForce encoder maps marker-based tactile images from heterogeneous sensors into force-grounded tactile tokens, which are concatenated with vision and language tokens for action decoding. The UniForce encoder remains frozen throughout. 

\begin{figure*}[t]
    \centering
    \includegraphics[width=0.8\linewidth]{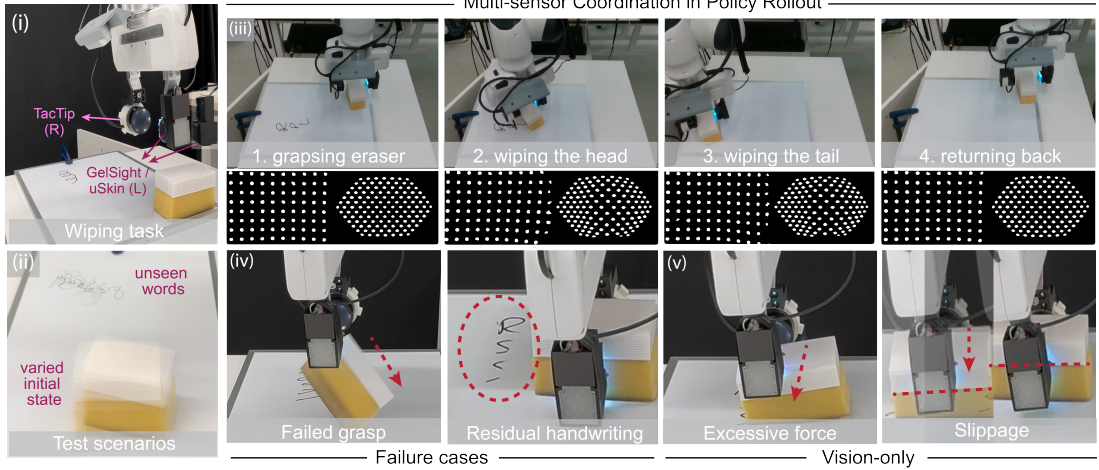}
    \caption{\textbf{Whiteboard wiping with heterogeneous tactile sensors.} (i) Setup for wiping task. (ii) Varied test scenarios. (iii) Policy rollout with multi-sensor coordination. (iv) Failure cases. (v) Inappropriate force results in slippage and over-pressing.
    }
    \vspace{-7pt}
    \label{fig:vtla_results}
\end{figure*}

\textbf{Data.}
We fix TacTip on the right finger and swap the left finger sensor between GelSight and uSkin (Fig.~\ref{fig:vtla_results}).
We collect 50 task trajectories with varied words and initial conditions.

\textbf{Comparison groups.}
\begin{itemize}
    \item \textbf{V:} Vision only ($\pi_{0.5}$ without tactile tokens). 
    \item \textbf{V+T:} Vision + TacTip. ($\pi_{0.5}$ with tactile tokens)
    \item \textbf{V+G:} Vision + GelSight (zero-shot transfer from V+T). 
    \item \textbf{V+U:} Vision + uSkin (zero-shot transfer from V+T).
    \item \textbf{V+T+G:} Vision + TacTip + GelSight (policy finetuned with both sensors; UniForce frozen).
    \item \textbf{V+T+U:} Vision + TacTip + uSkin (policy finetuned with both sensors; UniForce frozen).
\end{itemize}



\textbf{Metric.}
We report the success rate (SR) over 20 rollouts per setting.
A rollout is considered successful if it completes all task steps \emph{without} any failed grasp and leaves residual writing \emph{below} the threshold (residual area \(\le 10\%\)). 

\begin{figure}[t]
    \centering
    \includegraphics[width=1\linewidth]{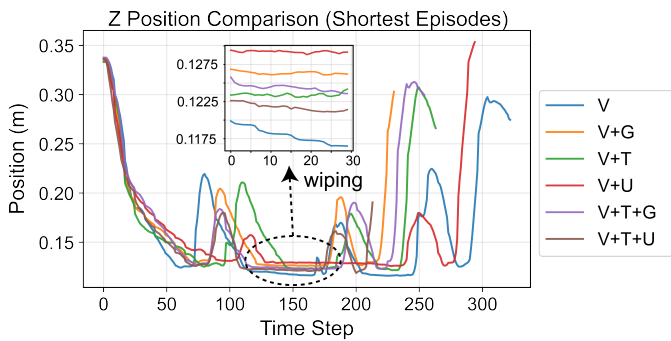}
    \caption{\textbf{Z-axis position of robot hand during wiping}. }
    \vspace{-7pt}
    \label{fig:z_axis}
\end{figure}

Table~\ref{tab:wiping_modalities} shows that adding tactile sensing improves SR over vision-only (V: 20\%).
UniForce further enables transferring a TacTip-trained policy (50\%) to GelSight (45\%) and uSkin (60\%) \emph{without} per-sensor tactile-encoder finetuning. Using tactile signals from both fingers (V+T+G: 75\% and V+T+U: 80\%) yields higher SR than using a single tactile sensor, suggesting that UniForce can exploit heterogeneous tactile inputs for force-aware skill learning. One plausible explanation is that bilateral tactile feedback provides complementary contact information (e.g., normal and shear cues on both sides), which helps maintain stable contact during wiping and reduces failures from slip or excessive force shown in Table~\ref{tab:wiping_modalities}. We further plot the end-effector Cartesian \(z\)-position for the shortest successful episodes in Fig.~\ref{fig:z_axis}.
The flatter profile during the wiping phase indicates more stable vertical contact regulation when using UniForce with tactile sensing, consistent with fewer instances of inappropriate force (e.g., slip-induced height changes or over-pressing). 

\begin{table}[t]
  \centering
  \caption{Policy performance for wiping with different input modalities.}
  \label{tab:wiping_modalities}
  \begin{tabular}{l c cc cc}
    \toprule
    \multirow{2}{*}{Modality} & \multirow{2}{*}{\shortstack{Success\\Rate}}
      & \multicolumn{2}{c}{\shortstack{Failure\\Case}}
      & \multicolumn{2}{c}{\shortstack{Inappro.\\Force}} \\
    \cmidrule(lr){3-4}\cmidrule(lr){5-6}
      & & \shortstack{Failed\\Grasp} & \shortstack{Residual\\Handwriting} & Slippage & \shortstack{Excess\\Force} \\
    \midrule
    V       & 4/20 (20\%)  & 12 & 4 & 3 & 3 \\
    V+T     & 10/20 (50\%) & 8  & 2 & 2 & 1 \\
    V+G     & 9/20 (45\%)  & 9  & 2 & 2 & 1 \\
    V+U     & 12/20 (60\%) & 7  & 1 & 1 & / \\
    V+T+G   & 15/20 (75\%) & 3  & 2 & 1 & / \\
    V+T+U   & 16/20 (80\%) & 2  & 2 & 1 & / \\
    \bottomrule
  \end{tabular}
\end{table}

\section{Limitations \& Discussions}
\label{sec:limitations}

\subsection{Quasi-Static Equilibrium Assumption}
UniForce relies on a quasi-static force-equilibrium prior when collecting paired data. In practice, this assumption can be violated by fast motions, slip, compliance in the gripper mechanism, or imperfect alignment/parallelism between the two fingertips and the indenter. When equilibrium is violated, the left/right interactions may no longer correspond to the same contact force. A practical direction is to automate data collection with closed-loop control to enforce quasi-static contact (e.g., velocity/acceleration limits) and to actively correct misalignment, thereby improving the quality of force-paired supervision.

\subsection{Material Modeling and Long-Term Dynamics}
UniForce conditions the decoder on the reference observation \(I_{\text{ref}}\) and injects the latent force to reconstruct \(I_{\text{cur}}\), but it does not explicitly model material properties. Soft elastomers can exhibit nonlinear and history-dependent behaviors (e.g., hysteresis, viscoelasticity, and delayed recovery), and sensors with similar marker patterns but different skin hardness may respond differently under the same loading.
A promising direction is to incorporate explicit material/state variables (e.g., hardness parameters) into the forward model, enabling better robustness to skin variations and long-term material drift.

\subsection{Generalization to Unseen Sensing Modalities}
UniForce is designed for tactile sensors whose raw signals can be converted into a marker/taxel image representation. Generalizing to substantially different modalities (e.g., whisker-based tactile sensing) remains open, especially when the signal geometry does not admit a natural 2D marker-image canonicalization. Future work may require a broader set of physics constraints beyond two-finger force equilibrium.

\subsection{Tactile Integration in Large Models}
In our VTLA experiments, we incorporate UniForce tactile tokens by concatenating them with vision and language tokens at the action decoder input. This simple fusion may not fully leverage tactile information for high-level reasoning.
Future work could explore structured fusion, e.g., integrating tactile cues into the VLM reasoning stack while also providing force-grounded tokens to the low-level action module, to better exploit complementary information across modalities.

\section{Conclusion}
\label{sec:conclusion}

We presented \textbf{UniForce}, a unified latent force representation learning framework for robot manipulation with heterogeneous tactile sensors. UniForce leverages force-equilibrium data to learn a shared latent force space without requiring ground-truth force labels during representation learning. The learned latent force vector enables zero-shot transfer for downstream force estimation and force-aware policy learning across diverse tactile sensors. It provides a step toward scalable, force-grounded tactile representations that reduce per-sensor retraining and facilitate deployment across embodiments.

\bibliographystyle{plainnat}
\bibliography{references}

\end{document}